\documentclass[conference]{IEEEtran}
\IEEEoverridecommandlockouts

\usepackage{xcolor,soul,framed} 

\colorlet{shadecolor}{yellow}
\usepackage[pdftex]{graphicx}
\graphicspath{{../pdf/}{../jpeg/}}
\DeclareGraphicsExtensions{.pdf,.jpeg,.png}

\usepackage[cmex10]{amsmath}
\usepackage{array}
\usepackage{mdwmath}
\usepackage{mdwtab}
\usepackage{eqparbox}
\usepackage{url}

\usepackage{amssymb}
\graphicspath{ {./images/} }
\usepackage{orcidlink}
\usepackage{breqn}
\usepackage{physics}

\usepackage{amsmath}

\usepackage{tikz} 

\usepackage{float}

\usepackage{algorithm} 
\usepackage{algpseudocode}

\usepackage{breqn}
\usepackage{physics}
\usepackage{subcaption}

 \usepackage{adjustbox}

 \makeatletter
\newcommand*\bigcdot{\mathpalette\bigcdot@{.5}}
\newcommand*\bigcdot@[2]{\mathbin{\vcenter{\hbox{\scalebox{#2}{$\m@th#1\bullet$}}}}}
\makeatother

\begin{document}

\title{Enhancing Power Quality Event Classification with AI Transformer Models}

\author{\IEEEauthorblockN{Ahmad M. Saber\IEEEauthorrefmark{1}$^,$\IEEEauthorrefmark{4},
Amr~Youssef\IEEEauthorrefmark{7},
Davor~Svetinovic\IEEEauthorrefmark{1},
Hatem Zeineldin\IEEEauthorrefmark{1},
Deepa~Kundur\IEEEauthorrefmark{4}, and
Ehab El-Saadany\IEEEauthorrefmark{1}}
\IEEEauthorblockA{\IEEEauthorrefmark{1}
Advanced Power and Energy Center, EECS Department, Khalifa University, Abu Dhabi, UAE
}
\IEEEauthorblockA{\IEEEauthorrefmark{4}Department of Electrical and Computer Engineering, University of Toronto, Toronto, ON, Canada
}
\IEEEauthorblockA{\IEEEauthorrefmark{7}Concordia Institute for Information Systems Engineering,
Montreal, Quebec, Canada
}
}

\maketitle

\begin{abstract}
Recently, there has been a growing interest in utilizing machine learning for accurate classification of power quality events (PQEs). However, most of these studies are performed assuming an ideal situation, while in reality, we can have measurement noise, DC offset, and variations in the voltage signal's amplitude and frequency. Building on the prior PQE classification works using deep learning, this paper proposes a deep-learning framework that leverages attention-enabled Transformers as a tool to accurately classify PQEs under the aforementioned considerations.  The proposed framework can operate directly on the voltage signals with no need for a separate feature extraction or calculation phase. Our results show that the proposed framework outperforms recently proposed learning-based techniques. It can accurately classify PQEs under the aforementioned conditions with an accuracy varying between  99.81\%--91.43\%  depending on the signal-to-noise ratio, DC offsets, and variations in the signal amplitude and frequency. 
\end{abstract}

\section{Introduction}

The pervasive use of power electronics in smart grids introduces challenges to electric power supply quality. Classifying Power Quality Events (PQEs) is crucial for maintaining reliable and efficient power systems, meeting regulatory standards, ensuring customer satisfaction, and supporting grid planning and operation. Poor power quality adversely affects end users, leading to disruptions and damage. Prompt classification of PQEs enables utilities to address issues, enhancing overall customer satisfaction and improving maintenance practices for improved reliability. Furthermore, classification aids in identifying events like harmonics, which contribute to energy losses, allowing optimization for enhanced energy efficiency and cost reduction.

Recently, there has been a growing interest in exploiting the power of learning-based techniques for the classification of PQEs. Convolutional Neural Networks (CNNs) have demonstrated effectiveness in conjunction with other modules \cite{16, 18, DeepPower, b, c, e, n20}. Recurrent Neural Networks (RNNs) are also employed in PQE classification \cite{15}. Mishra emphasizes the prevalence of Artificial Neural Networks (ANNs) for PQE classification \cite{n4}. Qui et al. integrate time and frequency domain information using multi-fusion CNN \cite{b}. Addressing hidden patterns, a deep autoencoder followed by feature categorization is proposed \cite{n14}. Sahani and Dash combine variational mode decomposition with an improved particle swarm optimization technique \cite{n15}. Fu et al. introduce PowerCog, utilizing optimized CNN and SVM for precise identification of PQEs in noisy environments \cite{e}. Gu et al. tackle the challenge of the simultaneous occurrence of multiple PQEs, introducing a label-guided attention method (LGAN) with a single dimension CNN \cite{a}. Chiam et al. present two methods for PQE classification \cite{chiam2021power,10010655}, yet they do not account for normal signal variations and DC offset in measurements. Additionally, a single-shot power quality disturbance detection approach based on CNN is proposed \cite{c}. Investigating adversarial attacks on CNN-based PQE classifiers, solutions are proposed \cite{n20}.

Compared to prior works, this paper presents a deep learning-based framework (DLF) that uses attention-enabled transformers to classify PQEs considering measurement noise and DC offset, as well as practical fluctuations in the voltage's magnitude and frequency. This novelty is achieved thanks to the advantages over CNN/RNN approaches tried previously.
The above sources of error greatly affect the performance of learning-based PQEs classifiers, as highlighted in some of the previous works \cite{b,d,PQconfpaper}, and therefore it is important to include all of them while developing a PQEs classifier.
The proposed transformer is directly trained on raw PQEs signals, with no need for a separate feature extraction phase. 
Additionally, the performance of the proposed solution is compared with other learning-based approaches.

Section II presents a preliminary PQEs classification problem. In Section III, a model-free Transformer-based framework for the classification of PQEs is developed. Simulation results are reported in Section IV. A sensitivity analysis is conducted in Section V. Finally, the paper concludes with Section VI.

\section{ Power Quality Events  }

Measured voltage signals in distribution systems exhibit different characteristics. They can be pure sinusoidal, or distorted by one or multiple PQ disturbances or by measurement errors. These signals can be modeled using a set of parametric equations. For instance, a pure sine wave can be mathematically represented as \cite{DeepPower}:

\begin{equation}
s(t)=A \textcolor{white}{.} sin( 2  \pi f t  - \theta)
\end{equation}

\noindent where $s$ is a signal that can represent current or voltage, $A$ is the signal amplitude, $f$ is the signal frequency, $t$ denotes the time, and $\theta$ denotes the possible phase shift.
Similarly, other PQEs, e.g. sag, swell, interruption, harmonics, impulsive transients, oscillatory transients, and flicker, can be expressed using modified versions of Equation (1) \cite{DeepPower,n20}. More details can be found in power system textbooks and references like \cite{DeepPower,n20}.
A total of 10 PQEs classes are simulated in this paper as depicted in Table \ref{Tab:PQEs}.

\begin{table}[t!]
\centering
\begingroup
\label{tab:1}
\caption{Simulated PQEs Classes}
\begin{tabular}{c| c| c| c  }\hline
\centering
\makebox{   Class   } &\makebox{   Description   } 
&\makebox{   Class   } &\makebox{     Description   }\\ \hline
C1 & Pure sinusoidal & C6 & Interruption   \\  
C2 &Sag	 & C7 & Impulsive transients   \\ 
C3 &  Swell & C8 &  Harmonics  \\ 
C4 & Oscillatory transients  & C9 &Sag + harmonics   \\
C5 & Flicker & C10 &Swell + harmonics  \\\hline
\end{tabular}
\label{Tab:PQEs} 
\endgroup
\end{table}

\section{Deep Learning-Based Framework for PQEs Classification Using Transformers}

Classification methods based on learning have emerged in diverse application domains, including PQEs classification.  In learning-based methodologies, the input data is processed through a multi-layered network, where each layer's output serves as the input for the subsequent layer \cite{saber2023VVC}. This facilitates the incorporation of multiple levels of data abstraction, which is very useful in PQEs classification 
\cite{  15, DeepPower}.
In this paper, we investigate the potential and benefits of deep transformers with attention layers \cite{Transformer} for accurate classification of PQEs considering measurement devices' imposed noise and DC offset, and variations in the voltage signal's amplitude and frequency, which can all occur in reality.
Transformers are an alternative to traditional RNN architectures, which have been used for PQEs classification. 
The main advantage of the Transformer over traditional sequence-based classifiers is that transformers can process the entire sequence, i.e., the entire voltage signal (sample) to be analyzed, in parallel, rather than processing it sequentially. This allows the Transformer to be more efficient and accurate, which is useful for PQEs classification under signal distortions that may not be recognized by focusing on only a small portion of the sequence.
Other advantages of Transformers over CNN/RNN approaches tried previously, include efficient capturing of long-range dependencies with the self-attention mechanism, effective handling of sequential data through positional encodings, crucial for tasks with sequences, scalability achieved via stacked layers for larger datasets and complex tasks, and enhanced global context understanding  \cite{Transformer, Transformer_Paper}. 

\subsection{Deep Transformers for PQEs Classification}

We use the Keras implementation of the transformers applied to time series classification \cite{Keras_Transformer}, which is based on the original transformers proposed in \cite{Transformer_Paper}.
Overall, the Transformer model represents a significant advance in the field of time series data processing and has achieved impressive results on a variety of tasks. The use of attention mechanisms allows the model to effectively process sequential data and generate accurate output, and the attention mechanism allows it to focus on multiple parts of the input simultaneously.
\begin{figure}[t!]
  \centering  \includegraphics[width=0.43\columnwidth]{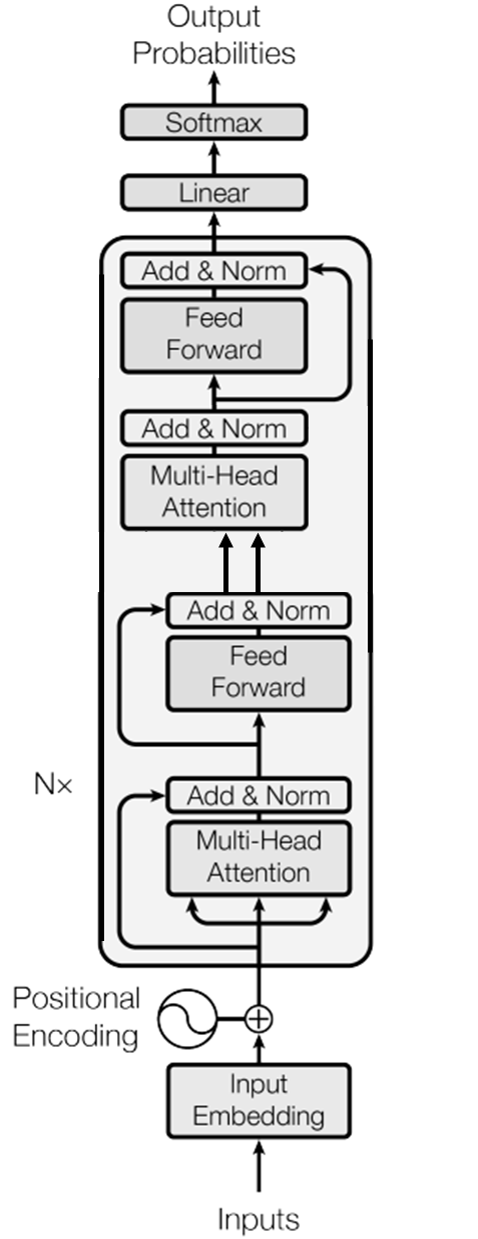}
  \centering
  \caption{Transformer Model Architecture \cite{Transformer_Paper}.}
  \centering
  \label{fig:Tmodel}
\end{figure}
Fig.  \ref{fig:Tmodel} shows an illustration of the Transformer architecture. The proposed Transformer is a neural network architecture that relies on self-attention mechanisms to process sequential data.
The Transformer model consists of embedding layers, multiple layers of attention and fully connected (FC) layers. 
The model processes the input time series data N times.
Firstly, the model takes in a sequence of input vectors $X = {x_1, x_2, ..., x_n}$ and produces a sequence of hidden states $H = {h_1, h_2, ..., h_n}$. Next, this sequence of hidden states as well as an additional sequence of target vectors $Y = {y_1, y_2, ..., y_m}$ are used to produce a final output sequence $O = {o_1, o_2, ..., o_m}$.
The attention mechanism used in the Transformer model can be represented by:

\begin{equation}
 Attention (Q, K, V) = f(\frac{QK^T}{\sqrt{d_k}})V
\end{equation}
 
\noindent in which $Q$, $K$, and $V$ are matrices representing the query, key, and value vectors, respectively, $d_k$ is the dimensionality of the keys, and  $f$ is the activation function often being Softmax \cite{15}.
$Q$ and $K$, which are derived from the input time series data, represent the information that the model is seeking at a particular time step and are used to compute the attention scores, determining the importance of different time steps in relation to each other. $V$  represents the voltage signal's content that the model should pay attention to, and is used to calculate the weighted sum based on the attention scores, which is then used as the input to the next layer of the transformer.
A key aspect of Transformers is the use of attention, which allows the model to attend to multiple positions in the input sequence simultaneously. This is achieved by using multiple attention heads, each of which processes a different part of the input sequence. The output of the attention heads is then concatenated and passed through a final feed-forward network before being used to predict the output as follows:

\begin{equation}
MultiHead(Q, K, V) = Concat(head_1, ..., head_h)W^O 
\end{equation}

\noindent where 

\begin{equation}
head_i = Attention(QW_i^Q, KW_i^K, VW_i^V)
\end{equation}

\noindent Herein, $W_i^Q$, $W_i^K$, $W_i^V$, and $W^O$ are weight matrices. Moreover, the feed-forward network:
\begin{equation}
FFN(x) = max(0, xW_1 + b_1)W_2 + b_2 
\end{equation}

\noindent where $x$ is the input, and $W_1$, $b_1$, $W_2$, and $b_2$ are weight matrices and biases, respectively.
The Transformer architecture includes residual connections and layer normalization, which help to improve the stability and performance of the model.

\section{Simulation Results}
 
\subsection{Performance Evaluation Metric}

The performance of the proposed framework is evaluated using the weighted average accuracy ($acc$)  computed as \cite{saber2022anomaly}

\begin{equation}
acc = \frac{\sum_{i=1}^{n} w_i \cdot acc_i}{\sum_{i=1}^{n} w_i}
\end{equation}

\noindent where  $n$ is the number of power quality classes, $w_i$ is the weight for class $i$, and the 
$acc_i$ is the accuracy for class  $i$ determined as

\begin{equation}
acc_i =  \frac{ tp_i + tn_i  }{ tp_i + tn_i + fp_i + fn_i  }  
\end{equation}

\noindent where the terms $tp$, $tn$, $fp$, and $fn$ represent true positives, true negatives, false positives, and false negatives, respectively$-$all calculated from the viewpoint of class $i$. That is, for class $i$, the $tp$ cases are correctly identified instances from the $i$th PQE class, while $tn$ cases are correctly classified instances that belong to the remaining $(n - i)$ classes.  For instance, $acc_1$, for the class C1 of pure sine waves,  the positive class is C1, while the remaining (C2 to C10) are treated as the negative class for the sake of calculating $acc_1$.
In the same example, a disturbance observation misclassified as a pure sine wave is a $fp$ case. Conversely, pure sine waves misclassified as a disturbance fall under the $fn$ cases.

\subsection{ Dataset Generation and Transformer Model Training }

Ten classes of PQEs are simulated, which are pure sinusoidal (no disturbance), sag, swell, temporary transients, flicker, interruption, impulsive transients, and harmonics, in addition to two simultaneous categories which are sag + harmonics, and swell + harmonics.
Additionally, measurement errors including additive white Gaussian noise of signal-to-noise ratios (SNRs) ranging from 5 to 60 dB, and DC offset, in the range of 0-10\%, are imposed on each generated case. 
Moreover, the frequency and amplitude of each signal are allowed to vary in the range of $\pm$ 5\% and $\pm$ 10\%, respectively, to account for normal variations in power systems. 
Table \ref{Tab:PQEs} presents a summary of the examined power quality events (PQEs). Two datasets are synthetically generated, in MATLAB R2023b environment, using each signal's parametric equation, with a sampling frequency of 10 kHz.
Dataset A is simulated without any measurement error and with fixed voltage amplitude and frequency (1pu), and Dataset B is simulated with noise and/or DC offset. With a total of 10 classes, each dataset has 100,000 signals and is balanced among the 10 classes. 
When calculating the $acc$ metric, all classes are given the same $w_i$ value, which is 1/$n$, where $n$ is the total number of classes.
Each signal was recorded for a duration of 10 cycles (equivalent to 0.2 seconds at 50Hz) and subsequently normalized. For each event, the corresponding feature values are determined. The simulated harmonics primarily included the 3rd, 5th, and 7th odd harmonics, which are the most frequently observed.
The generated dataset is then randomly shuffled and then divided into two main parts, one part for training and validation, and the other part for testing. 
\textcolor{black}{When splitting the datasets into training, validation, and testing, we ensured that in both the validation and test datasets, samples equally represent each class and also vary randomly within the same class, similar to the concept of stratified sampling.}
The training set includes 80\% of the samples, the validation set includes 10\% of the samples and the testing set includes the remaining 10\%. 
The generated dataset has a size of 7GB. Therefore, we utilized a setup with  12 NVIDIA GTX 1050 16-GB GPUs.
The Transformer model is trained
using a combination of supervised learning and self-supervised learning, where the model is trained 
to predict the next token in a sequence given the previous ones \cite{Transformer_Paper}. This allows the model to learn the structure and patterns of the input data, and improve its ability to generate coherent and accurate output.
In this research, the proposed transformer has 41 layers and  264,658 trainable parameters.
The utilized transformer has 4 blocks.
For simplicity, a single attention head is utilized with a size of 256. 
The 128 Multi-Layer Perceptron units have a dropout of 0.4 while the overall dropout is selected as 0.25. 
`Adam' optimizer is utilized with a learning rate of $10^{-4}$. 
The model is trained, over 100 epochs and with a batch size of 64, to optimize the sparse categorical accuracy. 
The early stopping criterion is forced with a patience of 10.  Training time was approximately 5 hours.

\subsection{ Techniques for Comparison}
 
 To establish a comparison, four DL techniques are also implemented.
 \textcolor{black}{The first is a convolutional neural network (CNN) model with  2 series groups of layers, and each group has a convolutional layer, a batch Normalization, and a max pooling layer, in addition to a flattening layer, a fully-connected layer with 16 nodes and the classification layer.
The second is a support vector machines (SVM) model with a quadratic kernel. The third is a Random Forest with min samples split set to 5. The fourth is a multi-layer perceptron (MLP) with 3 layers.}

 \subsection{Results}

\begin{figure*}[t!]
  \centering  
  \includegraphics[width=0.98\columnwidth]{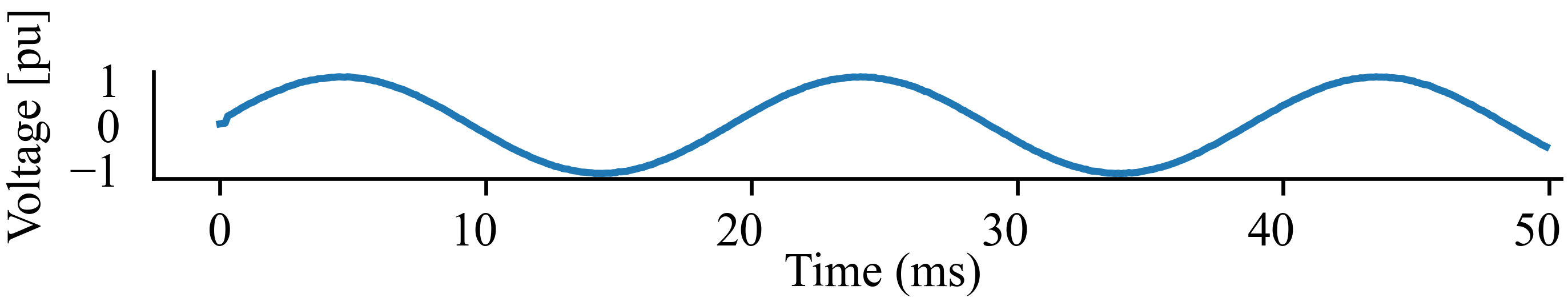} 
   \includegraphics[width=0.98\columnwidth]{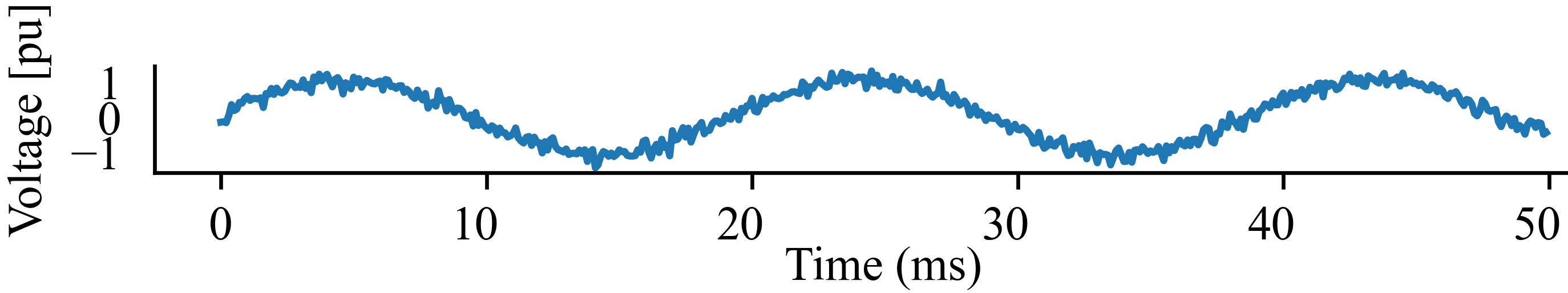}
  \\ \hspace{25pt} (a)  \hspace{225pt} (b) \\
 \includegraphics[width=0.98\columnwidth]{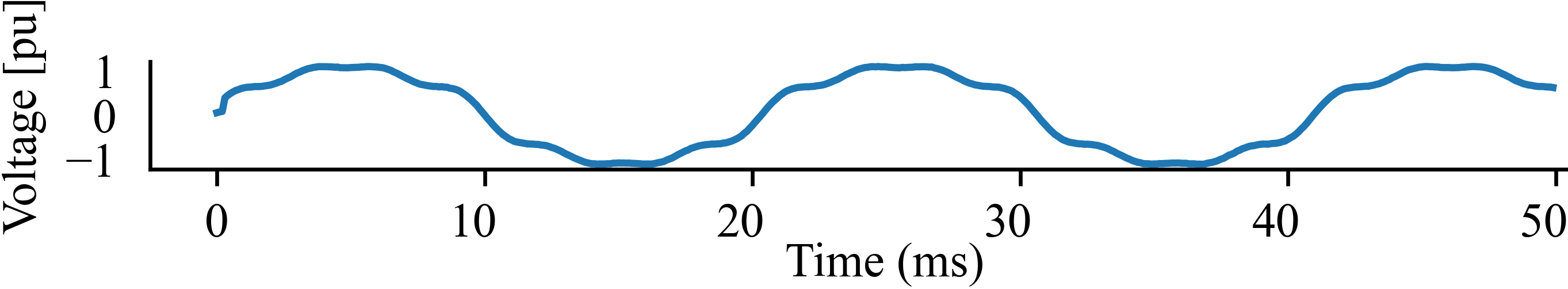}
  \includegraphics[width=0.98\columnwidth]{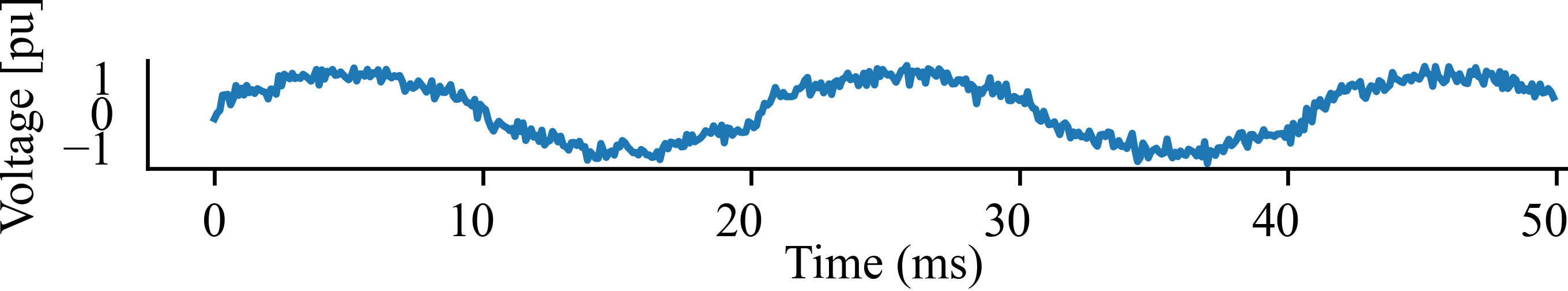}
  \\ \hspace{25pt} (c)  \hspace{225pt} (d) \\ 
 \includegraphics[width=0.98\columnwidth]{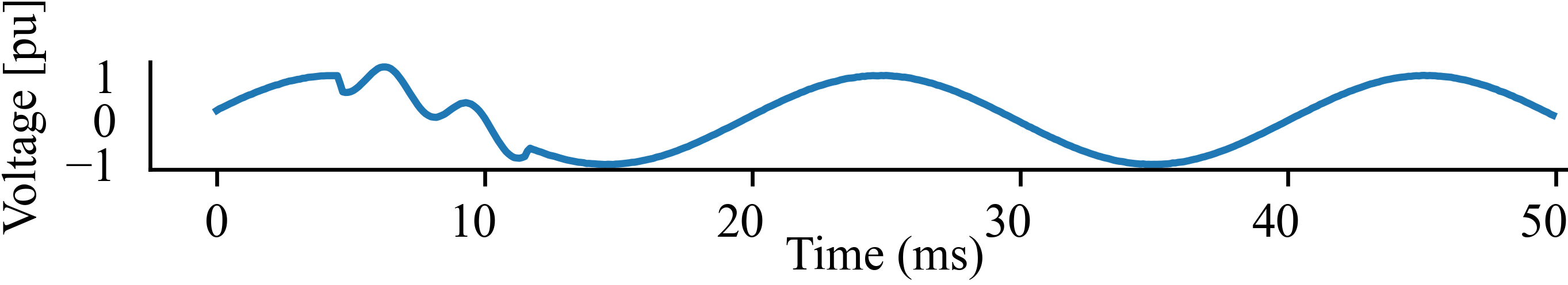}
  \includegraphics[width=0.98\columnwidth]{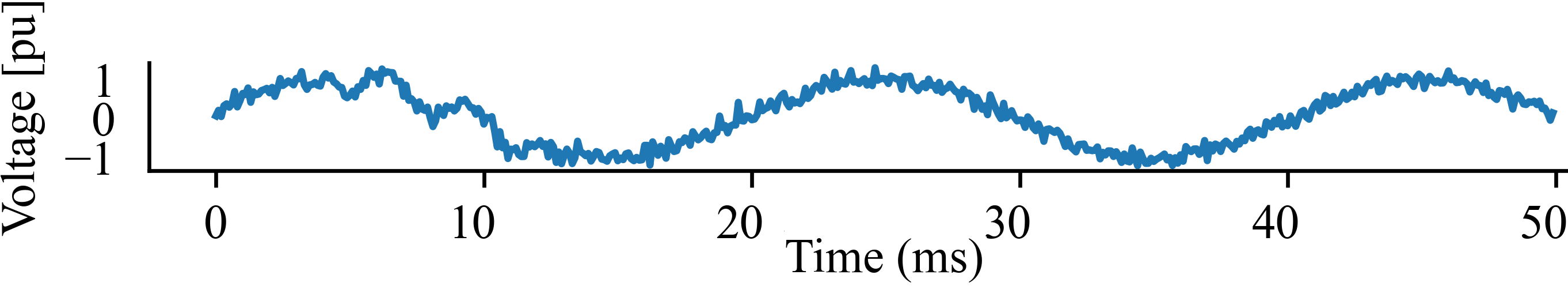}
  \\ \hspace{25pt} (e)  \hspace{225pt} (f) \\ 
  \centering
  \caption{PQEs Samples.}
  (a) pure sinusoidal,
  (b) pure sinusoidal + noise (SNR = 15dB), 
    (c) harmonics,
  (d) harmonics  + noise (SNR = 15dB),\\
     (e) temporary oscillatory transient,
  (f) temporary oscillatory transient  + noise (SNR = 15dB).
  \centering
  \label{fig:Harmonics_vs_pure_noise}
\end{figure*}

Firstly, the Transformer model is trained and tested on a dataset of PQEs that does not contain any noise or DC offset, whose voltage signal is exactly 50Hz, and whose initial amplitude ($A$) is 1pu. 
Overall, the utilized model achieves a testing $acc$  of  99.81\%, outperforming CNN, DNN, SVM, and Random Forests,  which resulted in 99.03\%, 98.6\%, 98.2\%, and 97.8\%, respectively as depicted in Table \ref{Tab:results}. 
The detailed confusion matrix is depicted in \ref{Tab:MatrixA}.
 Secondly, the proposed framework is tested under extreme conditions, with measurement noise whose SNR is as low as 5 dB, DC offset as high as 10\% and the voltage signal's frequency and amplitude deviate in the range of $\pm$ 5\% and $\pm$ 10\%, respectively. Our results indicate that, even under such conditions, the proposed Transformer approach achieves an $acc$ of 91.43\%, compared to CNN, DNN, Random Forests, and SVM,  which resulted in 88.9\%, 88.5\%, 88\%, and 86.7\%, respectively, as summarized in Table \ref{Tab:results2}. 
The confusion matrix for this scenario, depicted in Table \ref{Tab:MatrixB}, shows that classes C1 and C8, are the two most confused classes. This is because under the low-frequency measurement noise,  the time-domain representation of these two cases, on which the transformers work, is similar.
Low-frequency noise, with SNR greater than 35 dB, masks some of the signal's main features, causing them to be misclassified. 
A time-domain plot for a case of harmonics, a case of a pure sine wave contaminated with measurement noise, and a case of temporary transients are shown in Fig. \ref{fig:Harmonics_vs_pure_noise}.
However, using accurate measurement devices solves this problem.

 \begin{table}[t!]
\centering
\begin{minipage}{0.24\textwidth}  
\caption{Results part A
}
\begin{tabular}{c | c}\hline
\centering
\makebox{Algorithm} & \makebox{$acc$ } \\\hline
\textbf{Transformers} & \textbf{99.81\%} \\
CNNs & 99.03\% \\
DNNs & 98.6\% \\
SVMs & 98.2\% \\
Random Forests & 97.8\%  \\
\hline
\end{tabular}
 
 \vspace{2pt} Minimum SNR = 60 dB,\\  DC offset = 0, \\signal amplitude = 1pu, \\
 signal frequency = 50Hz
\label{Tab:results}
\end{minipage}
\hfill
\begin{minipage}{0.24\textwidth}  
\caption{Results part B
}
\begin{tabular}{c | c}\hline
\centering
\makebox{Algorithm} & \makebox{$acc$ } \\ \hline 
\textbf{Transformers} & \textbf{91.43\%}  \\
CNNs & 88.9\% \\
DNNs & 88.5\% \\
SVMs & 86.7\% \\
Random Forests & 88\% \\
\hline
\end{tabular}

\vspace{2pt} Minimum SNR = 5 dB,\\  DC offset = $\pm$ 10\%,\\ varying signal amplitude, \\
varying signal frequency
\label{Tab:results2}
\end{minipage}
\end{table}

\begin{table}[t!]
\centering
\caption{Confusion Matrix A}
\begin{adjustbox}{width= \columnwidth}
\begin{tabular}{c|cccccccccc}
\hline
\text{ } & C1 & C2 & C3 & C4 & C5 & C6 & C7 & C8 & C9 & C10 \\
\hline
C1 & 100 & 0 & 0 & 0 & 0 & 0 & 0 & 0 & 0 & 0 \\
C2 & 0 & 100 & 0 & 0 & 0 & 0 & 0 & 0 & 0 & 0 \\
C3 & 0 & 0 & 99.95 & 0 & 0 & 0 & 0.05 & 0 & 0 & 0 \\
C4 & 0.2 & 0 & 0 & 99.75 & 0 & 0 & 0 & 0.05 & 0 & 0 \\
C5 & 0.05 & 0 & 0.05 & 0.15 & 99.75 & 0 & 0 & 0 & 0 & 0 \\
C6 & 0 & 0 & 0 & 0 & 0 & 100 & 0 & 0 & 0 & 0 \\
C7 & 0 & 0 & 0 & 0 & 0 & 0 & 100 & 0 & 0 & 0 \\
C8 & 0 & 0 & 0 & 0.93 & 0 & 0 & 0 & 99.07 & 0 & 0 \\
C9 & 0 & 0 & 0 & 0 & 0 & 0 & 0.29 & 0 & 99.71 & 0 \\
C10 & 0 & 0 & 0 & 0.1 & 0 & 0 & 0 & 0 & 0 & 99.9 \\
\hline
\end{tabular}
\end{adjustbox}
 \label{Tab:MatrixA} 
\end{table}

\begin{table}[t!]
\centering
\caption{Confusion Matrix B}
\begin{adjustbox}{width=\columnwidth}
\begin{tabular}{c|cccccccccc}
\hline
\text{ } & C1 & C2 & C3 & C4 & C5 & C6 & C7 & C8 & C9 & C10 \\
\hline
C1 & 68.25 & 0 & 0.05 & 6.73 & 0.05 & 0 & 0.3 & 24.58 & 0 & 0.05 \\
C2 & 0 & 100 & 0 & 0 & 0 & 0 & 0 & 0 & 0 & 0 \\
C3 & 0 & 0 & 97.57 & 0.1 & 0.4 & 0 & 1.93 & 0 & 0 & 0 \\
C4 & 1.78 & 0 & 0 & 96.94 & 0 & 0.05 & 0.15 & 0.99 & 0.05 & 0.05 \\
C5 & 0.41 & 0 & 0.41 & 1.17 & 97.76 & 0 & 0 & 0.25 & 0 & 0 \\
C6 & 0.05 & 0 & 0 & 0.15 & 0 & 99.8 & 0 & 0 & 0 & 0 \\
C7 & 0.25 & 0 & 0.45 & 0.6 & 0 & 0 & 98.61 & 0 & 0.1 & 0 \\
C8 & 27.98 & 0 & 0.05 & 11.23 & 0 & 0.05 & 0.15 & 60.48 & 0 & 0.05 \\
C9 & 0.05 & 0.1 & 0 & 0.05 & 0 & 0 & 2.24 & 0 & 96.98 & 0.58 \\
C10 & 0.34 & 0.59 & 0 & 0.74 & 0 & 0 & 0 & 0.1 & 0.34 & 97.88 \\
\hline
\end{tabular}
\end{adjustbox}
 \label{Tab:MatrixB} 
\end{table}

\section{Hyper-Parameter Values Sensitivity Analysis }

\textcolor{black}{The results of implementing the proposed model on Dataset A are almost perfect compared to Dataset B which shows more variability and thus room for improvement. Therefore, an additional analysis on Dataset B is performed where we examine how changing hyper-parameters affects the solution's performance.}

\  \textit{Scenario 1}: A model is trained with smaller hyper-parameter values; a head size of  128, 2 transformer blocks, 64 MLP units, an MLP dropout of 0.2, and a general dropout of 0.12.  It is noticed that $acc$  did not significantly change, as the obtained  $acc$ in this scenario is 
90.93\%. 

\noindent\textit{ Scenario 2}: Smaller learning rates can result in improved $acc$  at the cost of increasing the training time, which is not a limitation in our problem. In this scenario, a smaller learning rate, of the value $2.5 \times 10 ^{-5}$, is utilized and the Transformer is retrained. The $acc$  is found to be   91.12\%.

\noindent \textit{Scenarios 3, 4, and 5}: Herein, the impact of increasing the number of attention heads on the model's performance is evaluated. Three new models, with 2, 3, and 4 attention heads, respectively, are trained and tested. 
Interestingly, increasing the number of heads resulted in little change to the model's performance, as the $acc$ in Scenarios 4, 5, and 6 are found to be  91.32\%,  91.41\%, and  91.16\%, respectively.

\noindent \textit{Scenario 6}: 
Stochastic gradient descent (SGD) \cite{SDG}, an alternative optimizer to `Adam', is utilized, and the Transformer model is re-trained. The $acc$  in this scenario is  84.18\%, reflecting that `Adam' is a better optimizer in this problem. Naturally, any improperly optimized model performs poorly.

The sensitivity analysis results are summarized in Table \ref{Tab:S_results}. 
\textcolor{black}{Across the 6 scenarios, the model consistently performed well despite minor variations in hyperparameter values.}

\begin{table}[t!]
\centering
\begingroup
\caption{Effect of Varying Transformers Hyper-Parameters }
\begin{tabular}{c | c| c| c}\hline
\centering
\makebox{   Transformer Model   } &\makebox{     $acc$          }  
& \makebox{   Transformer Model   } &\makebox{     $acc$          } 
\\\hline
Scenario 1	& 90.93\%    &  Scenario 4	&    91.41\%   \\
Scenario 2	&   91.12\%   &  Scenario 5	& 91.16\%    \\
 Scenario 3	&   91.32\%    &   Scenario 6	&   84.18\%   \\ 
\hline 
\end{tabular}

\vspace{2pt} Base-Case Transformer Model:   91.43\%  (Under measurement error and signal variations)
\label{Tab:S_results}
\endgroup
\end{table}

\section{Conclusion}

This paper explored the challenge of accurately classifying PQEs under measurement noise, DC offset, and variations in the voltage signal's amplitude and frequency, which are unavoidable conditions in practice. A deep-learning framework is proposed based on attention-enabled Transformer models to classify PQEs under the aforementioned conditions. Our results show that the proposed model outperforms other DL techniques, and thus can be implemented in future PQE classifiers.
While numerical simulations can provide valuable insights in controlled environments, enable the possibility of exploring extreme conditions, and provide detailed information, it is necessary to evaluate the proposed model for real-world implementation, commencing with hardware-in-the-loop tests similar to previous work \cite{saber2023cyber}.
Future work involves assessing the scheme's performance under noise, DC offset, and voltage variations together, addressing misclassifications under low-frequency noise, and exploring the scalability and resilience to large-scale systems and adversarial attacks.

\ifCLASSOPTIONcaptionsoff
  \newpage
\fi

 \bibliographystyle{IEEEtran} 
 \bibliography{output.bib}

\begin{thebibliography}{10}
\providecommand{\url}[1]{#1}
\csname url@samestyle\endcsname
\providecommand{\newblock}{\relax}
\providecommand{\bibinfo}[2]{#2}
\providecommand{\BIBentrySTDinterwordspacing}{\spaceskip=0pt\relax}
\providecommand{\BIBentryALTinterwordstretchfactor}{4}
\providecommand{\BIBentryALTinterwordspacing}{\spaceskip=\fontdimen2\font plus
\BIBentryALTinterwordstretchfactor\fontdimen3\font minus \fontdimen4\font\relax}
\providecommand{\BIBforeignlanguage}[2]{{%
\expandafter\ifx\csname l@#1\endcsname\relax
\typeout{** WARNING: IEEEtran.bst: No hyphenation pattern has been}%
\typeout{** loaded for the language `#1'. Using the pattern for}%
\typeout{** the default language instead.}%
\else
\language=\csname l@#1\endcsname
\fi
#2}}
\providecommand{\BIBdecl}{\relax}
\BIBdecl

\bibitem{16}
J.~Li, H.~Liu, D.~Wang, and T.~Bi, ``Classification of power quality disturbance based on s-transform and convolution neural network,'' \emph{Frontiers in Energy Research}, vol.~9, p. 708131, 2021.

\bibitem{18}
M.~A. Rodriguez, J.~F. Sotomonte, J.~Cifuentes, and M.~Bueno-L{\'o}pez, ``Power quality disturbance classification via deep convolutional auto-encoders and stacked lstm recurrent neural networks,'' in \emph{2020 International Conference on Smart Energy Systems and Technologies (SEST)}.\hskip 1em plus 0.5em minus 0.4em\relax IEEE, 2020, pp. 1--6.

\bibitem{DeepPower}
N.~Mohan, K.~Soman, and R.~Vinayakumar, ``Deep power: Deep learning architectures for power quality disturbances classification,'' in \emph{2017 international conference on technological advancements in power and energy (TAP Energy)}.\hskip 1em plus 0.5em minus 0.4em\relax IEEE, 2017, pp. 1--6.

\bibitem{b}
W.~Qiu, Q.~Tang, J.~Liu, and W.~Yao, ``An automatic identification framework for complex power quality disturbances based on multifusion convolutional neural network,'' \emph{IEEE Trans. Industr. Inform.}, vol.~16, no.~5, pp. 3233--3241, 2019.

\bibitem{c}
C.~Iturrino-Garc{\'\i}a, G.~Patrizi, A.~Bartolini, L.~Ciani, L.~Paolucci, A.~Luchetta, and F.~Grasso, ``An innovative single shot power quality disturbance detector algorithm,'' \emph{IEEE Trans. Instrum. Meas.}, vol.~71, pp. 1--10, 2022.

\bibitem{e}
L.~Fu, K.~Yan, and T.~Zhu, ``Powercog: A practical method for recognizing power quality disturbances accurately in a noisy environment,'' \emph{IEEE Trans. Industr. Inform.}, vol.~18, no.~5, pp. 3105--3113, 2021.

\bibitem{n20}
J.~Tian, B.~Wang, J.~Li, and Z.~Wang, ``Adversarial attacks and defense for cnn based power quality recognition in smart grid,'' \emph{IEEE Trans. Netw. Sci. Eng.}, vol.~9, no.~2, pp. 807--819, 2021.

\bibitem{15}
N.~K. Manaswi, N.~K. Manaswi, and S.~John, \emph{Deep learning with applications using python}.\hskip 1em plus 0.5em minus 0.4em\relax Springer, 2018.

\bibitem{n4}
M.~Mishra, ``Power quality disturbance detection and classification using signal processing and soft computing techniques: A comprehensive review,'' \emph{International Trans. electrical energy systems}, vol.~29, no.~8, p. e12008, 2019.

\bibitem{n14}
C.~Ge, R.~A. de~Oliveira, I.~Y.-H. Gu, and M.~H. Bollen, ``Deep feature clustering for seeking patterns in daily harmonic variations,'' \emph{IEEE Trans. Instrum. Meas.}, vol.~70, pp. 1--10, 2020.

\bibitem{n15}
M.~Sahani and P.~K. Dash, ``Fpga-based semisupervised multifusion rdcnn of process robust vmd data with online kernel rvfln for power quality events recognition,'' \emph{IEEE Trans. Neural Netw. Learn. Syst.}, vol.~33, no.~2, pp. 515--527, 2020.

\bibitem{a}
D.~Gu, Y.~Gao, Y.~Li, Y.~Zhu, and C.~Wu, ``A novel label-guided attention method for multilabel classification of multiple power quality disturbances,'' \emph{IEEE Trans. Industr. Inform.}, vol.~18, no.~7, pp. 4698--4706, 2021.

\bibitem{chiam2021power}
D.~H. Chiam and K.~H. Lim, ``Power quality disturbance classification using transformer network,'' in \emph{International Conference on Cyber Warfare, Security and Space Research}.\hskip 1em plus 0.5em minus 0.4em\relax Springer, 2021, pp. 272--282.

\bibitem{10010655}
D.~H. Chiam, K.~H. Lim, and K.~H. Law, ``Detection of power quality disturbances using wavelet-based convolutional transformer network,'' in \emph{2022 International Conference on Green Energy, Computing and Sustainable Technology (GECOST)}, 2022, pp. 150--154.

\bibitem{d}
R.~Machlev, M.~Perl, J.~Belikov, K.~Y. Levy, and Y.~Levron, ``Measuring explainability and trustworthiness of power quality disturbances classifiers using xai—explainable artificial intelligence,'' \emph{IEEE Trans. Industr. Inform.}, vol.~18, no.~8, pp. 5127--5137, 2021.

\bibitem{PQconfpaper}
A.~M. Saber, A.~Selim, V.~Kadkikar, H.~Zeineldin, and E.~El-Saadany, ``Fast deep-learning-based recognition of multiple power quality events under noise and dc offset,'' in \emph{2023 IEEE Conference on Power Electronics and Renewable Energy (CPERE)}.\hskip 1em plus 0.5em minus 0.4em\relax IEEE, 2023, pp. 1--6.

\bibitem{saber2023VVC}
A.~M. Saber, A.~Youssef, D.~Svetinovic, H.~Zeineldin, and E.~El-Saadany, ``Learning-based detection of malicious volt-var control parameters in smart inverters,'' in \emph{IECON 2023- 49th Annual Conference of the IEEE Industrial Electronics Society}, 2023, pp. 1--6.

\bibitem{Transformer}
J.~Vig, ``A multiscale visualization of attention in the transformer model,'' \emph{arXiv preprint arXiv:1906.05714}, 2019.

\bibitem{Transformer_Paper}
A.~Vaswani, N.~Shazeer, N.~Parmar, J.~Uszkoreit, L.~Jones, A.~N. Gomez, {\L}.~Kaiser, and I.~Polosukhin, ``Attention is all you need,'' \emph{Advances in neural information processing systems}, vol.~30, 2017.

\bibitem{Keras_Transformer}
T.~Ntakouris, ``Timeseries classification with a transformer model,'' \emph{keras.io/examples/timeseries/timeseries_classification_transformer/}, 2021.

\bibitem{saber2022anomaly}
A.~M. Saber, A.~Youssef, D.~Svetinovic, H.~H. Zeineldin, and E.~F. El-Saadany, ``Anomaly-based detection of cyberattacks on line current differential relays,'' \emph{IEEE Trans. Smart Grid}, vol.~13, no.~6, pp. 4787--4800, 2022.

\bibitem{SDG}
N.~S. Keskar and R.~Socher, ``Improving generalization performance by switching from adam to sgd,'' \emph{arXiv preprint arXiv:1712.07628}, 2017.

\bibitem{saber2023cyber}
A.~M. Saber, A.~Youssef, D.~Svetinovic, H.~Zeineldin, and E.~F. El-Saadany, ``Cyber-immune line current differential relays,'' \emph{IEEE Trans. Industr. Inform.}, 2023, doi: 10.1109/TII.2023.3310769.

\end{thebibliography}

\end{document}